\documentclass{bmvc2k}

\usepackage{amsmath}
\usepackage{amssymb}
\usepackage{mwe}
\usepackage{multirow}
\usepackage{makecell}
\usepackage[table]{xcolor}
\usepackage{xcolor}
\usepackage{color}
\usepackage{algorithm}
\usepackage{algpseudocode}
\usepackage{lipsum}

\title{Semi-MoE: Mixture-of-Experts meets Semi-\\Supervised Histopathology Segmentation}

\addauthor{Nguyen Lan Vi Vu*}{vi.vuvivu2203@hcmut.edu.vn}{1}
\addauthor{Thanh-Huy Nguyen*}
{thanhhun@cmu.andrew.edu}{2}
\addauthor{Hoang-Thien Nguyen}{n21dccn080@student.ptithcm.edu.vn}{3}
\addauthor{Daisuke Kihara}{dkihara@purdue.edu}{4}
\addauthor{Tianyang Wang}{tw2@uab.edu}{5}
\addauthor{Xingjian Li$^\dagger$}{lixj04@gmail.com}{2}
\addauthor{Min Xu$^\dagger$}{mxu1@cs.cmu.edu}{2}

\addinstitution{
 University of Technology,\\
 Ho Chi Minh City, Vietnam
}
\addinstitution{
 Carnegie Mellon University,\\
 Pennsylvania, USA
}
\addinstitution{
 Posts and Telecommunications \\Institute of Technology,\\
 Ho Chi Minh City, Vietnam
}
\addinstitution{
 Purdue University,\\
 Indiana, USA
}
\addinstitution{
 University of Alabama at Birmingham,\\
 Alabama, USA
}

\runninghead{Vi Vu \& Huy Nguyen, et al.}{Semi-MoE on Histopathology Segmentation}

\begin{document}

\maketitle



\begin{abstract}
Semi-supervised learning has been employed to alleviate the need for extensive labeled data for histopathology image segmentation, but existing methods struggle with noisy pseudo-labels due to ambiguous gland boundaries and morphological misclassification. This paper introduces Semi-MOE, to the best of our knowledge, the first multi-task Mixture-of-Experts framework for semi-supervised histopathology image segmentation. Our approach leverages three specialized expert networks: A main segmentation expert, a signed distance field regression expert, and a boundary prediction expert, each dedicated to capturing distinct morphological features. Subsequently, the Multi-Gating Pseudo-labeling module dynamically aggregates expert features, enabling a robust fuse-and-refine pseudo-labeling mechanism. Furthermore, to eliminate manual tuning while dynamically balancing multiple learning objectives, we propose an Adaptive Multi-Objective Loss. Extensive experiments on GlaS and CRAG benchmarks show that our method outperforms state-of-the-art approaches in low-label settings, highlighting the potential of MoE-based architectures in advancing semi-supervised segmentation. 
Our code is available at \url{https://github.com/vnlvi2k3/Semi-MoE}.



\end{abstract}

\section{Introduction}
Colon cancer is one of the deadliest malignancies, spreading to vital organs and ranking as the second leading cause of cancer-related deaths worldwide \cite{islam2022colon}. Histopathology image analysis is the gold standard for cancer diagnosis, particularly in assessing glandular morphology to determine the grade of colon cancer \cite{graham2019mild,gurcan2009histopathological}. In pathology image analysis, fully supervised learning models require accurate pixel-level annotation, which
can be time-consuming and require expert knowledge. Additionally, annotators may have different labels, affecting how the model is trained and assessed. Due to these constraints, deploying fully supervised models widely in real-world applications is difficult.

Recent semi-supervised semantic segmentation (SSSS) frameworks employ various strategies to enhance model performance under limited supervision. 
Among these approaches, teacher-student learning is a popular paradigm where a teacher network provides pseudo labels or soft targets to guide a student network to learn from unlabeled data. The teacher network is often an EMA (Exponential Moving Average) version of the student. Typical approaches include Mean Teacher \cite{tarvainen2017mean}, FixMatch \cite{fixmatch} and Uncertainty-Aware Mean Teacher \cite{yu2019uncertainty}. Co-training is another dominant SSSS framework where two or more peer networks perform mutual learning by exchanging pseudo labels. Recent approaches such as  Cross Pseudo Supervision (CPS) \cite{chen2021-CPS} and X-Net \cite{zhou2024xnet} empirically show more promising results on histopathology segmentation tasks. Both teacher-student learning and co-training are designed to propagate pseudo-supervision to the target network via auxiliary learners with diverse data views. Some other work focuses more on elaborated data augmentation strategies such as style transfer \cite{wang2020unlabeled},  multi-resolution consistency \cite{hsieh2024mscs}, generative models  \cite{guan2024global}, and frequency-domain augmentation with Wavelet transform \cite{zhou2024xnet}. Notably, TopoSemiSeg \cite{xu2023toposemiseg} follows the teacher-student framework and adopts topology-aware consistency regularization for histopathology images.



However, despite numerous advances in semi-supervised histopathology segmentation, some key challenges still exist and hence hinder effective learning from unlabeled examples. One major limitation in previous work is the \emph{homogeneity problem}. Specifically, for both teacher-student learning and co-training, the coordinated networks receive similar training data and loss signals. Even with the application of diverse data augmentation strategies, these learners still tend to converge to similar representations due to the strong learning capacity of Deep Neural Networks. As a result, errors made by one network are likely to be replicated rather than corrected by another homogeneous learner, increasing the risk of error accumulation over the training process. Moreover, previous works often overlook the inherent \emph{nature of histopathology data}, including intricate cancer structures, staining artifacts, and low foreground-background contrast. 
Therefore, the general SSSS approaches may not be optimal for representation learning on histopathology images. 

In this paper, we propose a multi-tasking Mixture-of-Experts framework \textbf{Semi-MoE} to overcome the limitations.
\textbf{Firstly}, Semi-MoE involves auxiliary tasks: Boundary map prediction \cite{oda2018besnet} and Signed Distance Function (SDF) \cite{xue2020shape} regression to enable robust extraction of both global tissue geometry and fine-grained features. As far as we know, our work is the first effort to 
design of a mixture of heterogeneous experts for semi-supervised medical segmentation, specifically on histopathology datasets.
\textbf{Secondly}, to enhance the reliability of pseudo labels generated by MoE, we design a novel Multi-Gate Pseudo-labeling (MGP) module that dynamically performs expert voting and refines feature representations for task-specific pseudo-label supervision.
\textbf{Thirdly}, we integrate multiple experts with an adaptive multi-objective loss that uses uncertainty-based weighting to balance task-specific losses and reduce the need for extensive manual hyperparameter tuning. \textbf{Finally}, our work achieves state-of-the-art performance on colon histopathology datasets (CRAG and GlaS), surpassing existing semi-supervised methods with extensive ablation studies in Dice and Jaccard.

\section{Related Work}
\textbf{Semi-Supervised Semantic Segmentation (SSSS).}
State-of-the-art SSSS approaches typically follow one of two primary strategies: Pseudo-Labeling \cite{yang2022ST++,fixmatch} and Consistency regularization \cite{chen2021-CPS,Ouali_2020_CVPR,tarvainen2017mean,Zhou_2023_ICCV}. Pseudo-labeling generates pseudo-labels for unlabeled samples using a model trained on the combination of the labeled samples and any previously pseudo-labeled samples \cite{cascante2021curriculum}. For example, FixMatch \cite{fixmatch} uses weak-to-strong augmentation consistency, while FlexMatch \cite{zhang2022flexmatchboostingsemisupervisedlearning} improves this by applying class-specific thresholds to better handle harder-to-learn classes. Consistency regularization originates from the objective of training models that remain invariant to various data augmentations \cite{fan2023revisiting}. For instance, \cite{Ouali_2020_CVPR} applied strong feature-space perturbations with multiple decoders to enforce prediction consistency, while \cite{jin2022semisupervised} enhanced the mean teacher model \cite{tarvainen2017mean} by adding a teacher assistant to improve knowledge transfer. In semi-supervised histopathology segmentation, XNet \cite{Zhou_2023_ICCV} enforces consistency between low- and high-frequency features obtained via wavelet transforms under a co-training \cite{chen2021-CPS, 9880095, 10203354} framework to enhance structural consistency. TopoSemiSeg \cite{xu2023toposemiseg} applies a noise-aware topological consistency loss via persistence diagrams to preserve structural signals in student predictions.

\textbf{Mixture of Experts in Computer Vision.}
The Mixture of Experts (MoE) framework, originally introduced in \cite{jacobs1993adaptive,jordan1994hierarchical}, is based on an intuitive yet powerful idea that different parts of a model, or experts, specialize in distinct tasks or aspects of the data. A gating network determines which experts to activate for a given input, enabling dynamic task allocation to the most relevant experts \cite{cai2025survey}. In the context of object detection, Mees et al. \cite{mees2016choosing} proposed a mixture of convolutional neural network (CNN) experts specialized for RGB, depth, and optical flow modalities, with a gating mechanism that adaptively fuses their outputs. Pavlitskaya et al. \cite{pavlitskaya2020using} first explored MoE for semantic segmentation, where each expert is trained on a semantic subset of the data. While earlier MoE approaches typically adopt a dense MoE with fixed-capacity structure across tasks, AdaMV-MoE \cite{chen2023adamv} introduces a sparsely activated architecture that dynamically adjusts the number of active experts per task, which provides stabilized multi-task learning and ultra-efficient inference. Yang et al. \cite{yang2024multi} propose MLoRE, a decoder-focused framework for multi-task dense prediction that enhances feature diversity and cross-task interaction by combining a global task-sharing convolution path with a Mixture-of-Low-Rank-Experts module. Nishi et al. \cite{nishi2024joint} introduce Joint-Task Regularization (JTR), a scalable regularization method that projects multi-task predictions and labels into a shared latent space to enforce consistency across tasks, under available partial labels.

\section{Methodology}
\begin{figure*}[t]
    \centering
    \includegraphics[width=1\linewidth]{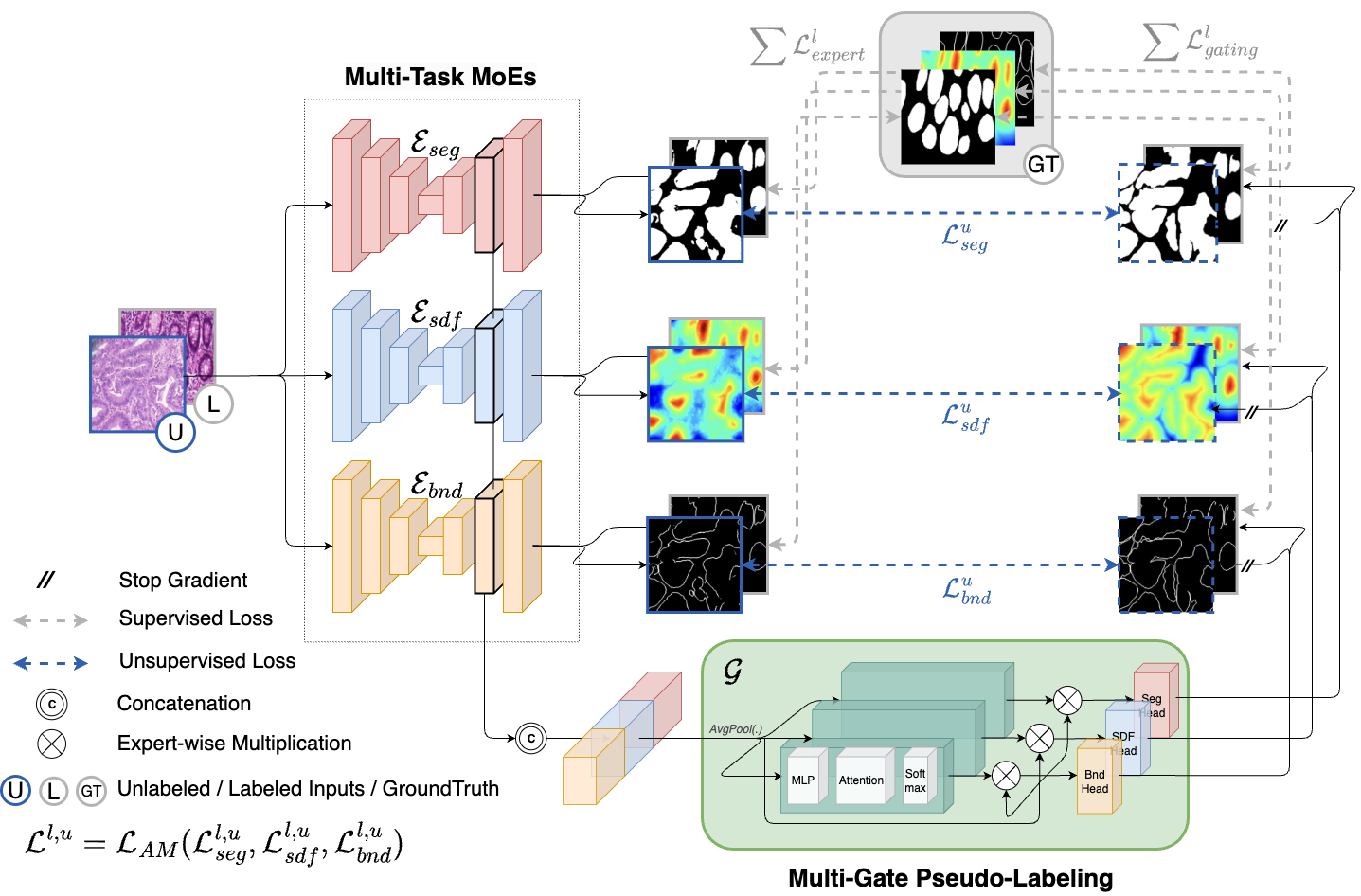}
    \caption{Overview of the proposed Semi-MoE framework.}
    \label{fig:main-figure}
\end{figure*}

In the context of semi-supervised segmentation, the training set consists of labeled samples $\mathbf{D}_l = \{(X_i^l, Y_i^l)\}_{i=1}^{N}$, and a larger unlabeled set $\mathbf{D}_u = \{(X_k^u)\}_{k=1}^{K}$ ($K \gg N$). Each input image $X_i^l, X_k^u$ is of size $3 \times H \times W$, with labeled samples $X_i^l$ being paired with ground-truth segmentation masks $Y_i^l$. Our workflow is illustrated in Fig. \ref{fig:main-figure}.

\subsection{Multi-task Mixture of Experts}

Boundary information is crucial for histopathology segmentation \cite{oda2018besnet}; however, intricate cancer structures, artifacts, and low foreground-background contrast often lead to noisy and ambiguous contours. As a result, model uncertainty is typically highest along gland boundaries. To address this, we introduce boundary map prediction as an auxiliary training task, where the boundary expert is penalized for boundary misclassification, thus strengthening the model’s ability to capture global tissue geometry and preserve topological information.

Despite this, two major segmentation errors arise due to morphological characteristics: false negatives occur when nuclei are misclassified as background due to their color similarity, while false positives result from stromal cells being mistakenly segmented as glandular structures due to staining artifacts. 
To mitigate these failures, we incorporate Signed Distance Function (SDF) regression as the second auxiliary task. Unlike binary masks, where local changes only affect nearby pixels, small variations in shape alter SDF values of multiple points globally \cite{xue2020shape}, thus encoding richer morphological information. Given the ground truth mask, the SDF for each pixel $x$ is defined as:
\begin{equation}
    SDF(x) =
\begin{cases} 
+d(x, \partial \Omega) , & \text{if } x \in \Omega_{in} \\ 
0, & \text{if } x \in \partial \Omega \\ 
-d(x, \partial \Omega), & \text{if } x \in \Omega_{out}, 
\end{cases}
\end{equation}
where $\partial \Omega$ denotes the boundary, $d(x, \partial \Omega)$ is the Euclidean distance from $x$ to the boundary, and the sign indicates whether 
$x$ is inside or outside the region. The SDF values are normalized to the range $[-1, 1]$. Through SDF estimation, the SDF expert enhances the spatial awareness of the model
, improving its robustness in nuclei and non-glandular regions.

We formalize our approach as follows. Each expert $\mathcal{E}_m$, where $ m \in \{\text{seg}, \text{sdf},\text{bnd}\}$, adopts a U-Net encoder-decoder architecture
, followed by a task-specific head $\mathcal{C}_m$ for pixel-wise prediction.
We also introduce $\mathcal{G}$, a multi-gating module (Section \ref{sec: mgp}) that integrates information across experts. The outputs of individual experts are computed as:
\begin{equation}\label{predict}
P^{u,l}_{m} = \mathcal{C}_{m}(\mathcal{E}_{m}(X^{u,l}))
\end{equation}
The multi-gating module aggregates and refines the expert outputs:
\begin{equation}\label{fuse}
\{ G_{\text{seg}}^{u,l}, G_{\text{sdf}}^{u,l}, G_{\text{bnd}}^{u,l} \} = \mathcal{G}(\mathcal{E}_{\text{seg}}(X^{u,l}) \oplus \mathcal{E}_{\text{sdf}}(X^{u,l}) \oplus \mathcal{E}_{\text{bnd}}(X^{u,l})),
\end{equation}
where $\oplus$ denotes concatenation operation. Given the binary ground truth \(Y_{\text{seg}}^l\), we derive the SDF labels \(Y_{\text{sdf}}^l\) using Distance Transform \cite{felzenszwalb2012distance} and the boundary labels \(Y_{\text{bnd}}^l\) using Morphological Boundary Extraction \cite{aquino2010detecting}. In supervised phase, the labels $Y^l$ guide both experts and gating module's outputs. Specifically, Dice loss is applied to segmentation and boundary predictions, while Mean Squared Error is used for signed distance maps:
\begin{equation}\label{sup}
    \begin{aligned}
        \mathcal{L}_{\text{sup}} = &\sum_{m \in \{\text{seg}, \text{bnd}\}} \{ \text{Dice}(\text{softmax}(P_m^l), Y_m^l) + \text{Dice}(\text{softmax}(G_m^l), Y_m^l) \} \\
        &+ \| \tanh(P_{\text{sdf}}^l) - Y_{\text{sdf}}^l \|_2^2 + \| \tanh(G_{\text{sdf}}^l) - Y_{\text{sdf}}^l \|_2^2
    \end{aligned}
\end{equation}
In the unsupervised phase, pseudo-labels are derived from the gating predictions (see Fig. \ref{fig:main-figure}, the three pseudo-labels following the stop-gradient operations):
\begin{equation}\label{pseudo}
\hat{Y}_{\text{seg}}^u = \operatorname{argmax}(G_{\text{seg}}^u), \quad
\hat{Y}_{\text{sdf}}^u = \tanh(G_{\text{sdf}}^u), \quad
\hat{Y}_{\text{bnd}}^u = \operatorname{argmax}(G_{\text{bnd}}^u)
\end{equation}
The unsupervised loss is defined as:
\begin{equation}\label{unsup}
    \begin{aligned}
        \mathcal{L}_{\text{unsup}} = &\sum_{m \in \{\text{seg}, \text{bnd}\}} \text{Dice}(\text{softmax}(P_m^u), \hat{Y}_m^u) + \|\tanh(P_{\text{sdf}}^u) - \hat{Y}_{\text{sdf}}^u \|_2^2
    \end{aligned}
\end{equation}
The final training objective is:
\begin{equation}\label{total_loss}
    \mathcal{L}_{\text{total}} = \mathcal{L}_{\text{sup}} + \lambda_{\text{unsup}} \mathcal{L}_{\text{unsup}}, 
\end{equation}
where $\lambda_{\text{unsup}}$ balances the supervised and unsupervised losses.

\subsection{Multi-gating Module for Robust Pseudo-labeling} \label{sec: mgp}
\begin{figure*}[ht]
    \centering
    \includegraphics[width=1\linewidth]{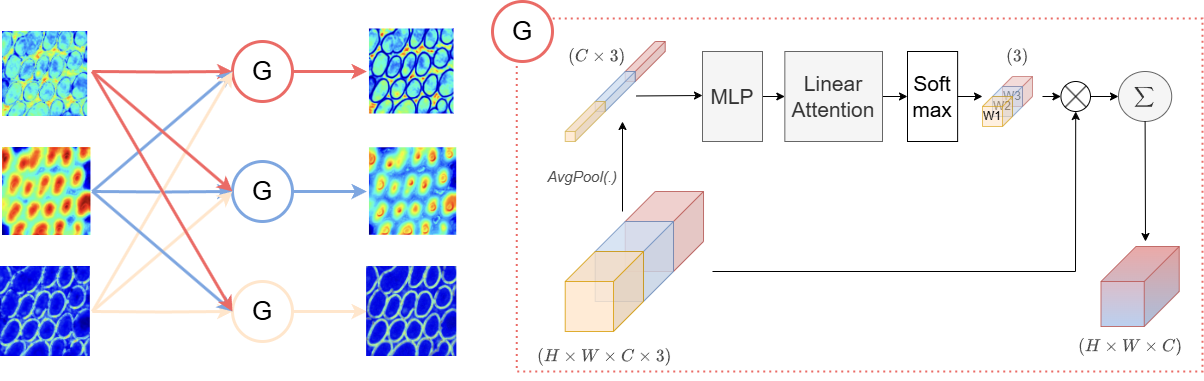}
    \caption{Multi-Gate Pseudo-Labeling (MGP) module. The left part illustrates how multi-gating networks dynamically aggregate features from task-specific experts, while the right part details the internal architecture of a gating network.}
    \label{fig:mgating}
\end{figure*}

Instead of naively using traditional Single-Gate Mixture-of-Experts \cite{royer2023revisiting} to generate pseudo labels for one task and transform them to supervise other tasks, we introduce the Multi-Gate Pseudo-labeling (MGP) module. Inspired by the complementary and interdependent nature of expert representations, we employ MGP with three separate gating sub-networks to dynamically fuse and refine feature representations from these task-specific experts. Unlike directly generating pseudo labels from prediction maps as in previous semi-supervised methods \cite{chen2021-CPS,tarvainen2017mean}, MGP enables gating networks to facilitate implicit voting among experts, allowing them to self-adjust their output representations, thus creating more robust pseudo-label with negligible parameter overhead.

As shown in Fig \ref{fig:mgating}, we concatenate expert features into $X^g = \bigoplus_{m} \mathcal{E}_{m}(X^{u,l}) \in \mathbb{R}^{M \times C \times H \times W}$, where $M$, $C$, $H \times W$ represent the number of experts, channels, and spatial dimensions, respectively. We first apply average pooling along the spatial dimension before computing the gating weights for each expert:
\begin{equation}
    W_m = \text{Softmax}\big(\text{LinearAttn}\big(\text{MLP}\big(\text{AvgPool}(X^g)\big)\big)\big),
\end{equation}
where $m \in \{\text{seg}, \text{sdf}, \text{bnd}\}$ and MLP is a fully connected layer with ReLU activation. The 
LinearAttn module consists of a fully connected layer for dimensionality reduction, dropout for regularization, and another linear projection to map the features to expert dimensions before calculating softmax-based probability distribution over experts. The weighted feature fusion is then performed as:

\begin{equation}
    G_m = {Head}_m(\sum_{i=1}^{3} W_{m,i} \otimes X^g_{i}),
\end{equation}
where $\otimes$ denotes expert-wise multiplication. 
$Head_m$ is used to project back the fused outputs to the task-specific output space. 
It mitigates negative transfer from weaker experts to stronger ones while leveraging shared knowledge to uplift underperforming experts.

\subsection{Adaptive Multi-objective Loss}
A common strategy for multi-objective optimization is weighting task losses, but performance is highly sensitive to these weights. Especially when task losses differ in scale and convergence rates, manual tuning becomes computationally expensive and prone to suboptimal solutions at each step. A multi-task loss can be formulated by maximizing the Gaussian likelihood under the assumption of homoscedastic uncertainty \cite{kendall2018multi}. Homoscedastic uncertainty does not depend on input data; instead, it varies across different tasks. Hence, it is called task-dependent uncertainty. In the context of multi-task MoEs, task uncertainty serves to quantify the relative confidence between tasks. Given an input $x$ and a neural network parameterized by weights $\theta$, let $f_{\theta}(x)$ denote the output, and $\sigma$ is the task-specific uncertainty. In our framework, the model jointly learns two classification tasks - semantic segmentation and boundary detection - and one regression task for SDF regression. The joint loss $\mathcal{L}(\theta, \sigma_{reg}, \sigma_{cls})$ is derived by minimizing the negative log-likelihood:
\begin{align}\label{uncertain_loss}
\mathcal{L}(\theta, \sigma_{reg}, \sigma_{cls})
&= -\log p(y_{reg}, y_{cls} = c | \theta) \notag \\
&= -\log \mathcal{N}(y_{reg}; f_{\theta}(x), \sigma_{reg}^2) \cdot \text{Softmax}(y_{cls} = c; f_{\theta}(x), \sigma_{cls}) \notag \\
&\approx \frac{1}{2\sigma_{reg}^2} \mathcal{L}_{reg}(\theta) + \frac{1}{\sigma_{cls}^2} \mathcal{L}_{cls}(\theta) 
+ \log \sigma_{reg} + \log \sigma_{cls},
\end{align}       
where $\mathcal{L}_{reg}(\theta) = \| y_{reg} - f_{\theta}(x) \|_2^2$ represents the Euclidean loss for SDF regression, while $\mathcal{L}_{cls}(\theta) = -\log \text{Softmax}(y, f_{\theta}(x))$ is the cross-entropy loss for segmentation and boundary prediction. A detailed derivation of Eq.~\eqref{uncertain_loss} is provided in the \textit{Appendix}. 

In practice, we model each uncertainty parameter $\sigma$ as a learnable, unconstrained scalar, which can cause instability in the original regularization term $\log \sigma$. To address this, we replace the original formulation with a numerically stable surrogate based on exponential functions. Since $e^{-\sigma}$ retains the monotonicity and positivity of $\frac{1}{\sigma^2}$, it provides a stable approximation for inverse variance weighting. Meanwhile, $e^{\sigma}$ acts as a smooth convex regularizer analogous to $\log \sigma$, penalizing overly large uncertainty. We thus define the Adaptive Multi-objective Loss applied to both supervised $\mathcal{L}_{\text{sup}}$ and unsupervised $\mathcal{L}_{\text{unsup}}$ loss terms, replacing their linear summation in Eq. \ref{sup} and \ref{unsup}:

\begin{equation}\label{am_loss}
\mathcal{L}_{AM} = e^{-\sigma_{\text{seg}}} \mathcal{L}_{\text{seg}} 
+ e^{-\sigma_{\text{sdf}}} \mathcal{L}_{\text{sdf}} 
+ e^{-\sigma_{\text{bnd}}} \mathcal{L}_{\text{bnd}} 
+ \gamma \sum_{j} e^{\sigma_j},
\end{equation}
where $\sigma_m$ is learnable parameter capturing task uncertainty, the regularization term $\gamma \sum_{j} e^{\sigma_j}$ prevents trivial solutions, and $\gamma$ is the only tunable hyperparameter in the objective function. The factor $\frac{1}2{}$ in $\frac{1}{\sigma_{reg}^2}$ is omitted as it can be absorbed into the learnable $\sigma_{sdf}$ without affecting the optimization. The entire training process is summarized in Algorithm~\ref{alg:semi-moe}.

\begin{algorithm}[H]
\caption{Training Semi-MoE}
\label{alg:semi-moe}
\begin{algorithmic}[1]
\Require Labeled set $\mathbf{D}_l=\{(X_i^l,Y_i^l)\}_{i=1}^N$, unlabeled set $\mathbf{D}_u=\{X_k^u\}_{k=1}^K$; 
experts $\{\mathcal{E}_m,\mathcal{C}_m\}_{m \in \{\text{seg},\text{sdf},\text{bnd}\}}$ with task uncertainty parameters $\{\sigma_m\}$; multi-gating module $\mathcal{G}$.
\Ensure Trained Semi-MoE model.
\For{each training iteration}
    \State Sample mini-batches $X^l \sim \mathbf{D}_l$, $X^u \sim \mathbf{D}_u$.
    \State Obtain expert predictions $P_m^{u,l} = \mathcal{C}_m(\mathcal{E}_m(X^{u,l}))$.
    \State Fuse expert representations via multi-gating: $G_m^{u,l} = \mathcal{G}(\oplus_m \mathcal{E}_m(X^{u,l}))$.
    \If{$X \in \mathbf{D}_l$}  \Comment{Supervised phase}
        \State Compute $\mathcal{L}_\text{sup}$ between $P_m^l$ and $Y^l_m$ (Eq.~\ref{sup}), task loss weighted by $\{\sigma_m\}$ (Eq.~\ref{am_loss}).
        \State Update parameters of $\{\mathcal{E}_m, \mathcal{C}_m, \mathcal{G}, \sigma_m\}$ via backpropagation.
    \EndIf
    \If{$X \in \mathbf{D}_u$}  \Comment{Unsupervised phase}
        \State Derive pseudo-labels $\hat{Y}_m^u$ from $G^u$.
        \State Compute $\mathcal{L}_\text{unsup}$ between $P_m^u$ and $\hat{Y}_m^u$ (Eq.~\ref{unsup}), task loss weighted by $\{\sigma_m\}$ (Eq.~\ref{am_loss}).
        \State Update parameters of $\{\mathcal{E}_m, \mathcal{C}_m, \sigma_m\}$ via backpropagation.
    \EndIf
\EndFor
\end{algorithmic}
\end{algorithm}

\section{Experiments and Results}

\subsection{Dataset and Evaluation Metrics}
The proposed method is comprehensively evaluated across two Colon Histology Image datasets: \textbf{CRAG} \cite{graham2019mild}  includes 213 annotated images, with 173 for training and 40 for testing, most of which have a resolution of 1512×1516. Gland segmentation is particularly challenging due to the significant variability in glandular morphology.\textbf{ GlaS} \cite{sirinukunwattana2017gland} is tailored for gland segmentation in colorectal adenocarcinoma and captures images from various stages of cancer progression. It provides 85 annotated training images (37 benign and 48 malignant) and 80 test images, with resolutions commonly around 775×522 or 589×453. Performance results of all models are trained with 10\% and 20\% labeled images in both datasets, averaged over 3-fold cross-validation. All experiments are evaluated using Dice Similarity Coefficient (DSC) and Jaccard (JC).
\subsection{Implementation Details}
We conduct all experiments on a single NVIDIA RTX 3090 Ti GPU, using Python 3.11, PyTorch 2.3, and CUDA 12.2. We train our model (built on backbone U-Net \cite{ronneberger2015u}) for 200 epochs to achieve convergence. The input images are first resized to 256×256. For the CRAG dataset, we apply random crop to 224×224 and resize back. All images are then augmented with random flips, rotations, and transpositions. We use SGD (momentum 0.9, weight decay $5 \times 10^{-5}$, initial LR 0.5) with a batch size of 2 yields the best results, $\gamma$ is set to 0.4 for GlaS, 0.8 for CRAG, and $\lambda_{unsup}$ is modeled as a time-dependent warming up function, reaching a max of 5.0. 

\subsection{Comparative Experiments}

\textbf{Quantitative Evaluation.} We benchmark Semi-MoE against nine state-of-the-art semi-supervised methods \cite{tarvainen2017mean,vu2019advent,yu2018pu,Ouali_2020_CVPR,chen2021-CPS,luo2022semi,luo2022semia,Zhou_2023_ICCV,xu2023toposemiseg}. 
For all methods except TopoSemiSeg\footnote{Due to reproduction challenges, we use reported numbers in the original paper \cite{xu2023toposemiseg}. Our experimental setup is the same as TopoSemiSeg.}, we run the experiments three times and calculate the mean performance and standard deviation. 
Table \ref{tab:comparison} highlights the superior performance of our framework, even with an extremely small amount of labeled samples. In contrast, many baseline methods experience a significant performance drop when the labeled data is reduced to 10\%. Moreover, unlike advanced methods such as CPS and TopoSemiSeg which demonstrate promising results only on specific datasets, our Semi-MoE consistently outperforms across all experimental groups.




\textbf{Qualitative Comparison.} Fig.~\ref{fig:comparison} showcases segmentation results on CRAG and GlaS under different labeling ratios. For CRAG (top two rows), Semi-MoE effectively preserves morphology, reducing fragmented predictions. Competing approaches struggle with gland boundaries, often misclassifying stromal cells as glands (false positives) or overlooking nuclei regions (false negatives). For GlaS (bottom two rows), Semi-MOE produces the most coherent glandular structures, accurately capturing the intricate boundaries with minimal artifacts. In contrast, baseline methods suffer from discontinuities and excessive noise, especially on the 10\% setting, highlighting Semi-MoE’s ability to preserve histological details. 

\begin{table*}[ht]
\caption{Quantitative experiments on GlaS and CRAG datasets. \textbf{Bold} and \underline{underline} denote the best and second-best performance.}
\label{tab:comparison}
\vspace{0.5mm}
    \centering
    \scriptsize
    \renewcommand{\arraystretch}{1.05}
    \setlength{\tabcolsep}{7pt} 
    \arrayrulecolor{gray} 

    \resizebox{1\textwidth}{!}{
    \begin{tabular}{c|l|cc|cc}
        \hline  
        \multirow{2}{*}{\textbf{Data}} & {\textbf{Labeled ratio}} & \multicolumn{2}{c|}{\textbf{20\%}} & \multicolumn{2}{c}{\textbf{10\%}} \\ 
        \cline{2-6}
        & {\textbf{Method}} & DSC (\%) $\uparrow$ & Jaccard (\%) $\uparrow$ & DSC (\%) $\uparrow$ & Jaccard (\%) $\uparrow$ \\
        \hline 
        \multirow{11}{*}{GlaS} & MT \cite{tarvainen2017mean} & $89.46 \pm 0.83$ & $80.94 \pm 1.35$ &$87.65 \pm 0.78$ & $78.02 \pm 1.23$ \\
        & EM \cite{vu2019advent} & $88.53 \pm 0.34$ & $79.42 \pm 0.56$ & $83.64 \pm 0.56$ & $71.89 \pm 0.79$ \\
        & UA-MT \cite{yu2018pu} & $89.27 \pm 0.12$ & $80.62 \pm 0.19$ &  $84.41 \pm 0.56$ & $73.03 \pm 0.83$ \\
        & CCT \cite{Ouali_2020_CVPR} & $89.49 \pm 0.09$ & $80.98 \pm 0.15$ &  $85.46 \pm 0.42$ & $74.51 \pm 0.64$ \\
        & CPS \cite{chen2021-CPS} & $\underline{90.25 \pm 0.53}$ & $\underline{82.23 \pm 0.92}$ & $\underline{89.15 \pm 0.44}$ & $\underline{80.43 \pm 0.31}$\\
        & URPC \cite{luo2022semi} & $88.92 \pm 0.45$ & $80.05 \pm 0.74$ & $81.59 \pm 0.62$ & $68.9 \pm 0.88$ \\
        & CT \cite{luo2022semia} &$88.45 \pm 0.3$ & $79.29 \pm 0.49$ & $81.65 \pm 0.61$ & $68.99 \pm 0.87$ \\
        & XNet \cite{Zhou_2023_ICCV} & $88.68 \pm 0.42$ & $79.67 \pm 0.67$ & $84.44 \pm 0.5$ & $73.07 \pm 0.75$ \\
        & TopoSemiSeg \cite{xu2023toposemiseg} & $89.5 \pm 0.0$ & $81.8 \pm 0.0$ & $87.8 \pm 0.0$ & $79.7 \pm 0.0$ \\ 
        
        \rowcolor{gray!20}
        & 
        \textbf{Semi-MoE (Ours)} & $\mathbf{90.4 \pm 0.32}$ & $\mathbf{82.48 \pm 0.51}$ & $\mathbf{89.23 \pm 0.41}$ & $\mathbf{80.55 \pm 0.45}$ \\
        \hline 
        \multirow{11}{*}{CRAG} & MT \cite{tarvainen2017mean} & $89.77 \pm 0.1$ & $81.43 \pm 0.15$ & $86.82 \pm 0.09$ & $76.71 \pm 0.14$ \\
         & EM \cite{vu2019advent} & $87.55 \pm 0.5$ & $77.93 \pm 0.8$ & $84.13 \pm 0.78$ & $72.61 \pm 1.17$ \\
        & UA-MT \cite{yu2018pu} & $88.84 \pm 0.11$ & $80.01 \pm 0.06$ & $83.97 \pm 0.46$ & $72.75 \pm 0.6$ \\
        & CCT \cite{Ouali_2020_CVPR} & $89.5 \pm 0.12$ & $80.99 \pm 0.19$ & $85.28 \pm 0.12$ & $74.34 \pm 0.18$ \\
        & CPS \cite{chen2021-CPS} & $89.42 \pm 0.15$ & $80.87 \pm 0.25$ & $85.52 \pm 0.31$ & $74.7 \pm 0.47$ \\
        & URPC \cite{luo2022semi} & $88.75 \pm 0.73$ & $79.25 \pm 0.25$ & $82.5 \pm 0.31$ & $72.1 \pm 0.61$ \\
        & CT \cite{luo2022semia} & $86.96 \pm 1.19$ & $75.9 \pm 0.33$ & $80.33 \pm 0.68$ & $67.13 \pm 0.95$  \\
        & XNet \cite{Zhou_2023_ICCV} & $88.1 \pm 0.15$ & $79.1 \pm 0.29$ & $86.67 \pm 0.05$ & $76.48 \pm 0.09$ \\
        & TopoSemiSeg \cite{xu2023toposemiseg} & $\underline{89.8 \pm 0.0}$ & $\underline{82.0 \pm 0.0}$ & $\underline{88.4 \pm 0.0}$ & $\underline{79.8 \pm 0.0}$ \\
        
        \rowcolor{gray!20} 
        &
        \textbf{Semi-MoE (Ours)} 
        & $\mathbf{91.06  \pm 0.08}$ & $\mathbf{83.59  \pm 0.11}$ & $\mathbf{89.27 \pm 0.12}$ & $\mathbf{80.61 \pm 0.18}$\\
        \hline 
    \end{tabular}
    }
\end{table*}

\textbf{Ablation study on multi-tasking experts.} We conducted an ablation study to investigate the impact of each expert on Semi-MoE. As shown in Table \ref{tab:ablation}, 
when incorporating only a single auxiliary expert, using Bnd outperforms Sdf, especially in the 10\% labeled setting. This suggests that Sdf is harder to learn with limited labels due to its dependence on precise per-pixel distance encoding. However, incorporating Sdf alongside Seg$+$Bnd provides consistent improvements, demonstrating its complementary benefits. It also shows the MGP module's effectiveness in dynamically refining expert knowledge for robust pseudo-labeling.

\begin{figure}[t]
    \centering
    \renewcommand{\arraystretch}{1.5}
    \setlength{\tabcolsep}{2pt}
    \begin{tabular}{c c c c c c c c c}
        & Image & GT & MT & CCT & CPS & URPC & XNet & Ours \\
        \rotatebox{90}{20\%} &
        \includegraphics[width=0.105\textwidth]{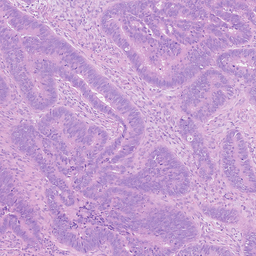} &
        \includegraphics[width=0.105\textwidth]{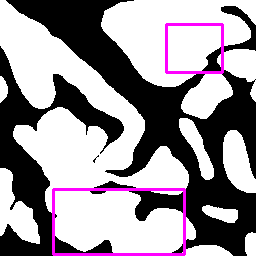} &
        \includegraphics[width=0.105\textwidth]{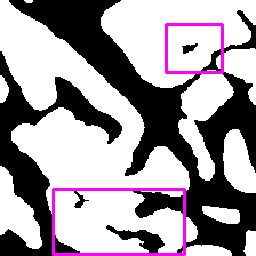} &
        \includegraphics[width=0.105\textwidth]{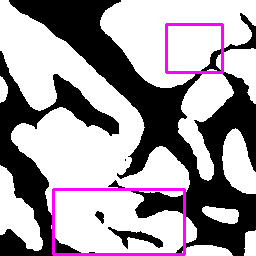} &
        \includegraphics[width=0.105\textwidth]{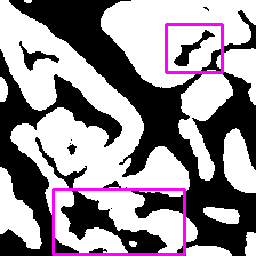} &
        \includegraphics[width=0.105\textwidth]{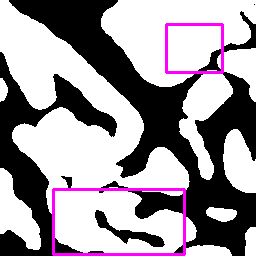} &
        \includegraphics[width=0.105\textwidth]{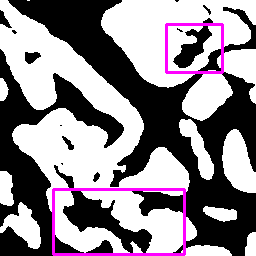} &
        \includegraphics[width=0.105\textwidth]{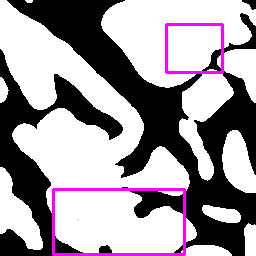} \\
        
        \rotatebox{90}{10\%} &
        \includegraphics[width=0.105\textwidth]{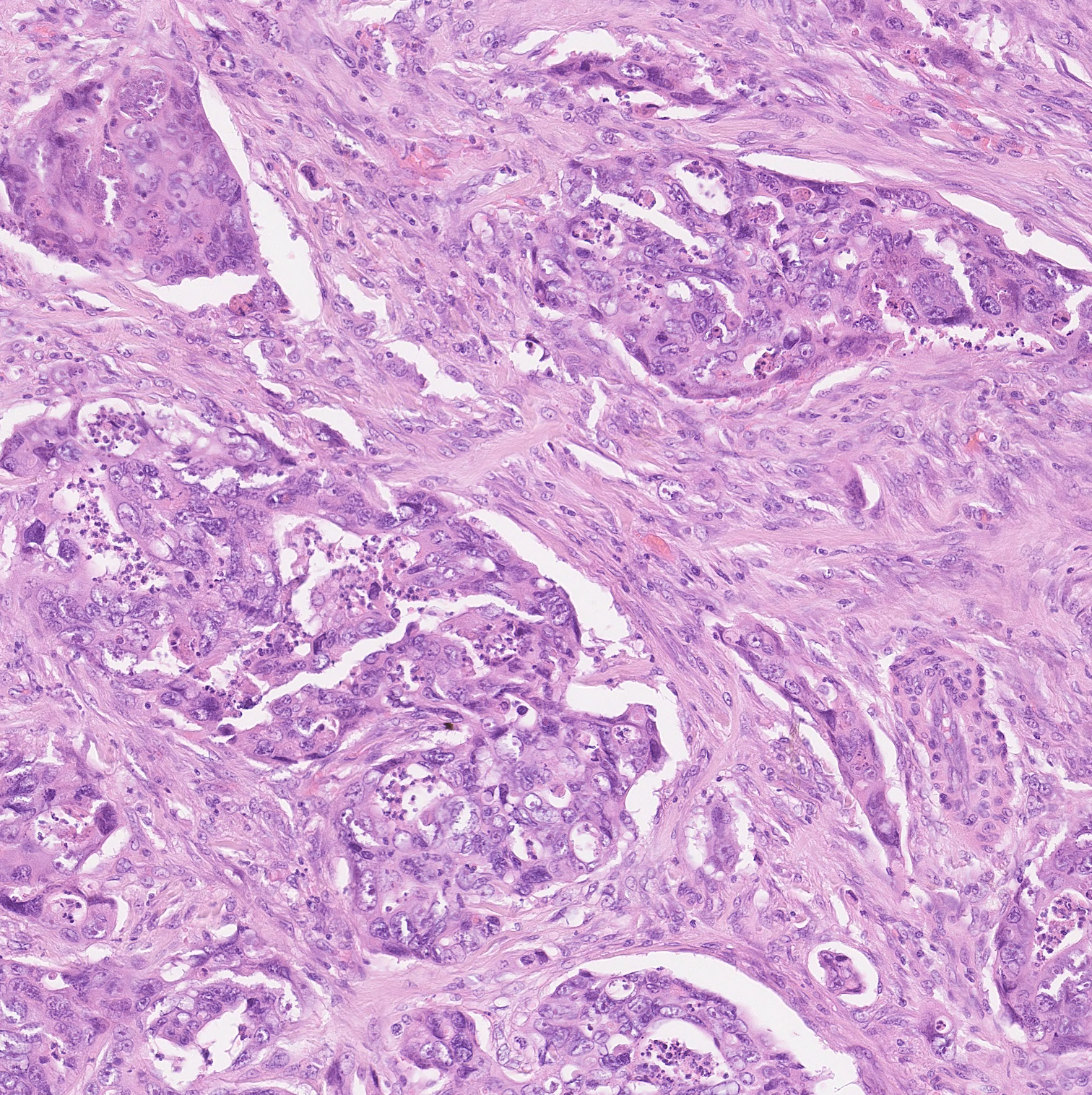} &
        \includegraphics[width=0.105\textwidth]{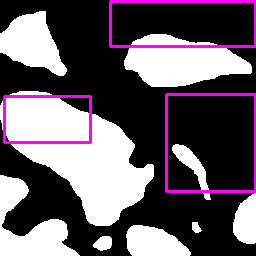} &
        \includegraphics[width=0.105\textwidth]{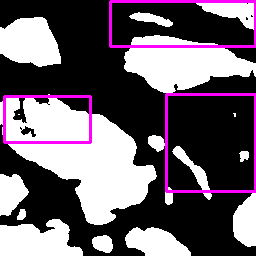} &
        \includegraphics[width=0.105\textwidth]{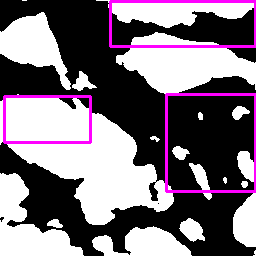} &
        \includegraphics[width=0.105\textwidth]{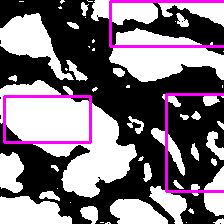} &
        \includegraphics[width=0.105\textwidth]{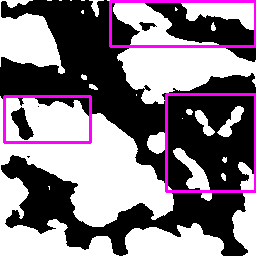} &
        \includegraphics[width=0.105\textwidth]{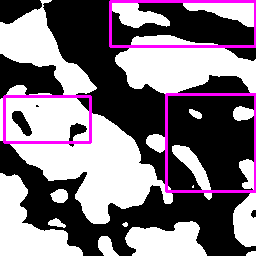} &
        \includegraphics[width=0.105\textwidth]{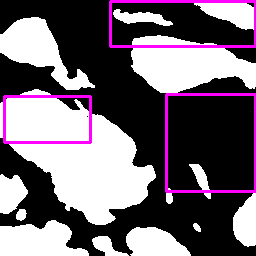} \\

        \rotatebox{90}{20\%} &
        \includegraphics[width=0.105\textwidth]{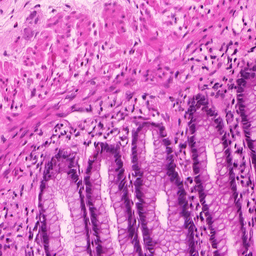} &
        \includegraphics[width=0.105\textwidth]{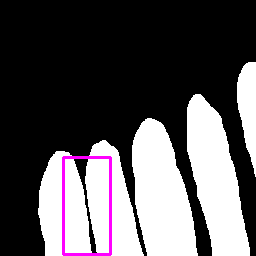} &
        \includegraphics[width=0.105\textwidth]{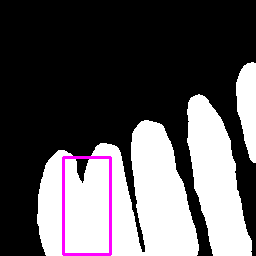} &
        \includegraphics[width=0.105\textwidth]{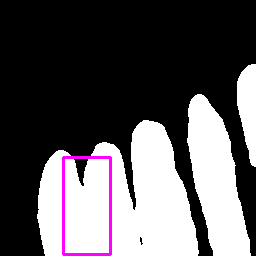} &
        \includegraphics[width=0.105\textwidth]{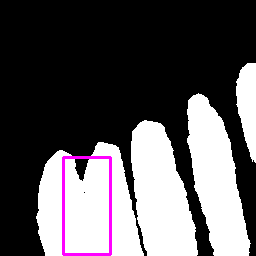} &
        \includegraphics[width=0.105\textwidth]{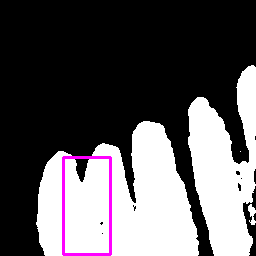} &
        \includegraphics[width=0.105\textwidth]{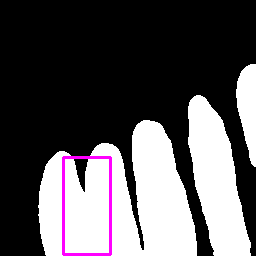} &
        \includegraphics[width=0.105\textwidth]{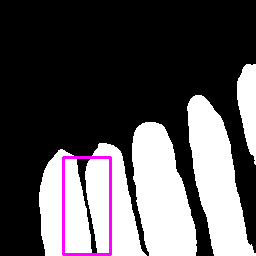} \\

        \rotatebox{90}{10\%} &
        \includegraphics[width=0.105\textwidth]{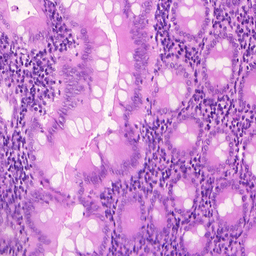} &
        \includegraphics[width=0.105\textwidth]{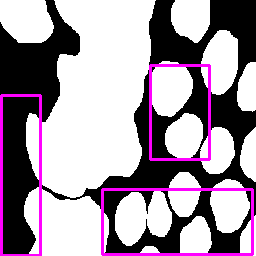} &
        \includegraphics[width=0.105\textwidth]{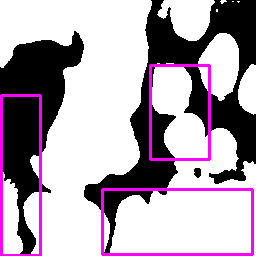} &
        \includegraphics[width=0.105\textwidth]{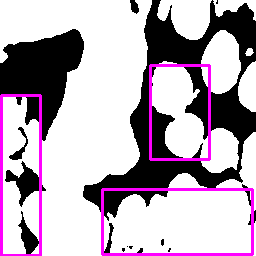} &
        \includegraphics[width=0.105\textwidth]{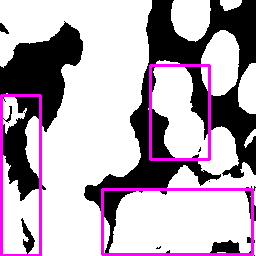} &
        \includegraphics[width=0.105\textwidth]{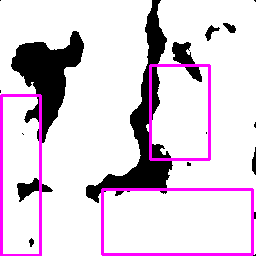} &
        \includegraphics[width=0.105\textwidth]{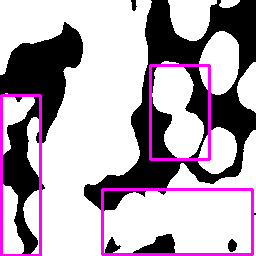} &
        \includegraphics[width=0.105\textwidth]{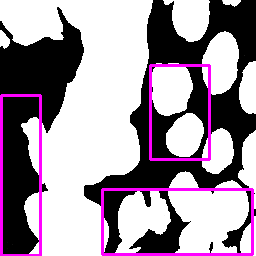} 
    \end{tabular}
    \caption{Qualitative results on CRAG (top two rows) and GlaS (bottom two rows).}
    \label{fig:comparison}
\end{figure}

\begin{table*}[t]
    \centering
    \centering
    \scriptsize
    \renewcommand{\arraystretch}{1.1}
    \setlength{\tabcolsep}{10pt}
    \caption{Ablation study on the impact of task-specific experts (seg is the main task).}
    \resizebox{1\textwidth}{!}{
        \begin{tabular}{c|c|c|c|cc|cc}
            \hline 

            Expert & \multirow{2}{*}{Seg} & \multirow{2}{*}{Sdf} & \multirow{2}{*}{Bnd} & \multicolumn{2}{c|}{20\%} & \multicolumn{2}{c}{10\%} \\
            \cline{0-0} \cline{5-8}
            Data &  &  &  & DSC & JC & DSC& JC  \\
            \hline 
             & \checkmark & \checkmark & & $87.93$ & $78.46$ & $85.34$ & $74.43$ \\
               GlaS  & \checkmark & & \checkmark & $\underline{89.04}$ & $\underline{80.25}$ & $\underline{88.81}$ & $\underline{79.87}$ \\
                 & \checkmark & \checkmark & \checkmark & $\textbf{90.4}$ & $\textbf{82.48}$ & $\textbf{89.23}$ & $\textbf{80.55}$ \\
            \hline 
             & \checkmark & \checkmark & & $\underline{90.73}$ &$ \underline{83.03}$ & $85.85$ & $75.21$ \\
              CRAG   & \checkmark & & \checkmark & $90.71$ &$ 83.00$ & $\underline{88.37}$ & $\underline{79.16}$ \\
                 & \checkmark & \checkmark & \checkmark & $\textbf{91.06}$ & $\textbf{83.59}$ & $\textbf{89.27}$ & $\textbf{80.61}$ \\
            \hline 
        \end{tabular}}
        
        \label{tab:ablation}
\end{table*}

\begin{table*}[ht]
    \centering
    \begin{minipage}{0.5\textwidth}
        \centering
        \scriptsize
        \renewcommand{\arraystretch}{1.2}
        \caption{Computation cost of methods.}
        \resizebox{0.85\textwidth}{!}{
        \begin{tabular}{ccc}
            \hline 
            Method & Params & Memory (GB)\\
            \hline
            MT & 155M & 4.58G  \\
            URPC & 116M & 3.62G   \\
            CCT & 133M & 4.11G \\
            CPS & 272M & 15.71G \\
            X-Net & 326M & 15.27G \\
            \rowcolor{gray!20} Semi-MoE & 181M & 12.36G \\
            
            \hline
        \end{tabular}}
        
        \label{table:computationcost}
    \end{minipage}
    \hfill
    \begin{minipage}{0.48\textwidth}  
        \centering
        \scriptsize
        \renewcommand{\arraystretch}{1.2}
        \caption{Ablation study for MGP and $\mathcal{L}_{AM}$ with 10\% labeled data.}
        \vspace{1mm}
        \resizebox{1\textwidth}{!}{
        \begin{tabular}{cccc}
            \hline 
            Data & Metric & w/o MGP & w/o $\mathcal{L}_{AM}$\\
            \hline
            Glas & DSC & -3.32 & -1.24 \\
            ~ & JC & -5.11 & -2.0 \\
            \hline
            CRAG & DSC & -0.22 & -1.91 \\
            ~ & JC & -0.34 & -3.06 \\
            
            \hline
        \end{tabular}}
        
        \label{table:ablation_more}
    \end{minipage}
    
\end{table*}

\textbf{Discussion on Computational Cost, Complexity, and Parameters.} As shown in Table \ref{table:computationcost}, although Semi-MoE introduces multiple task-specific experts and gating modules, it remains computationally efficient. The model uses 181M parameters and 12.36 GB of GPU memory, which is significantly lower than CPS (272M, 15.71G) and X-Net (326M, 15.27G). This efficiency is achieved through a shared U-Net encoder and lightweight gating heads that add minimal overhead. Compared to X-Net, Semi-MoE reduces parameter count by 44\% and memory usage by 19\%, while delivering superior segmentation performance. This highlights Semi-MoE’s strong balance between architectural complexity and scalability, making it practical for deployment in memory-constrained clinical environments.

\textbf{Ablation study on other components.} We evaluate our MGP against a Single-Gate variant, where only segmentation pseudo-labels are generated, while boundary and SDF labels are inferred via \cite{aquino2010detecting,felzenszwalb2012distance}. For $\mathcal{L}_{AM}$, we replace it with a linear combination of individual loss terms. Results in Table \ref{table:ablation_more} show that both MGP and $\mathcal{L}_{AM}$ are essential components. 

\textbf{Supplementary Materials.} These materials provide a detailed theoretical insight into the proposed Adaptive Multi-objective Loss, an extended ablation study based on Table~\ref{tab:ablation} with an additional task, and an investigation into the sensitivity of performance with respect to the hyperparameter $\gamma$.

\section{Conclusion}
In this paper, we introduce Semi-MoE - the first multi-task Mixture-of-Experts framework for semi-supervised histopathology image segmentation. By incorporating boundary map prediction and signed distance function regression as auxiliary tasks, we address ambiguous gland boundaries and morphological misclassification. Our Multi-Gating Pseudo-labeling module enables dynamic expert collaboration for robust pseudo-label fusion, and the novel Adaptive Multi-Objective loss further stabilizes the training process. Semi-MoE achieves state-of-the-art performance on histopathology semi-supervised learning benchmarks, highlighting Mixture-of-Experts as a promising paradigm for advancing semi-supervised segmentation.

\section{Acknowledgement}
This work was supported in part by U.S. NSF grant IIS-2211597.\\
We thank AI VIETNAM for facilitating GPU resources to conduct experiments.

\bibliography{egbib}
\end{document}